%% file: main.tex
\documentclass{article} %

\usepackage[accepted]{icml2019}

\input{math_commands.tex}

\usepackage{times}
\usepackage{relsize}
\usepackage[T1]{fontenc}
\usepackage[scaled=0.90]{inconsolata}
\usepackage[latin1]{inputenc}
\usepackage[english]{babel}

\usepackage{epsfig}
\usepackage{graphicx}
\usepackage{wrapfig}
\usepackage[belowskip=0pt,aboveskip=0pt,font=small]{caption}
\usepackage[belowskip=0pt,aboveskip=0pt,font=small]{subcaption}
\usepackage{amsmath, amsthm, amssymb, mathabx}
\usepackage{textcomp}
\usepackage{stmaryrd}
\SetSymbolFont{stmry}{bold}{U}{stmry}{m}{n}
\usepackage{upgreek}
\usepackage{bm}
\usepackage{cases}
\usepackage{mathtools}
\usepackage{arydshln}
\usepackage{multirow}

\definecolor{bluebell}{RGB}{52,31,151}
\definecolor{amour}{RGB}{238,82,83}
\usepackage[pagebackref=true,breaklinks=true,colorlinks,bookmarks=false,citecolor=bluebell,linkcolor=amour]{hyperref}

\usepackage{multirow}
\usepackage{rotating}
\usepackage{booktabs}

\usepackage{enumitem}
\usepackage[olditem,oldenum]{paralist}

\usepackage{alltt}
\usepackage{listings}

\usepackage{mysymbols}

\usepackage{url}
\usepackage{xspace}
\usepackage{comment}
\usepackage{color}
\usepackage{afterpage}
\usepackage{pdfpages}
\usepackage{framed}
\usepackage{fancybox}
\usepackage{cuted}
\usepackage{caption}

\usepackage{quoting}
\quotingsetup{vskip=0pt}
\usepackage{epigraph}

\definecolor{pumpkin}{RGB}{211,84,0}
\definecolor{ForestGreen}{RGB}{34,139,34}

\icmltitlerunning{TarMAC: Targeted Multi-Agent Communication}

\begin{document}

\twocolumn[
\icmltitle{TarMAC: Targeted Multi-Agent Communication}

\icmlsetsymbol{fairintern}{$\star$}

\vspace{-0.1in}
\begin{icmlauthorlist}
\icmlauthor{Abhishek Das}{gt,fairintern}
\icmlauthor{Th\'eophile Gervet}{mg}
\icmlauthor{Joshua Romoff}{mg,fair}\\
\vspace{0.1in}
\icmlauthor{Dhruv Batra}{gt,fair}
\icmlauthor{Devi Parikh}{gt,fair}
\icmlauthor{Michael Rabbat}{mg,fair}
\icmlauthor{Joelle Pineau}{mg,fair}
\end{icmlauthorlist}

\icmlaffiliation{gt}{Georgia Tech}
\icmlaffiliation{mg}{McGill University}
\icmlaffiliation{fair}{Facebook AI Research}

\icmlcorrespondingauthor{Abhishek Das}{{\tt abhshkdz@gatech.edu}}

\icmlkeywords{multi agent, reinforcement learning, communication}

\vskip 0.25in
]

\printAffiliationsAndNotice{}

\begin{abstract}
We propose a targeted communication architecture for multi-agent reinforcement learning,
where agents learn both \emph{what} messages to send and \emph{whom} to address them to
while performing cooperative tasks in partially-observable environments.
This targeting behavior is learnt solely from downstream task-specific reward
without any communication supervision.
We additionally augment this with a multi-round communication approach where agents
coordinate via multiple rounds of communication before taking actions in the environment.
We evaluate our approach on a diverse set of cooperative multi-agent tasks, of varying difficulties,
with varying number of agents, in a variety of environments ranging from $2$D grid
layouts of shapes and simulated traffic junctions to $3$D indoor environments,
and demonstrate the benefits of targeted and multi-round communication.
Moreover, we show that the targeted communication strategies
learned by agents are interpretable and intuitive.
Finally, we show that our architecture can be easily extended
to mixed and competitive environments, leading to improved performance
and sample complexity over recent state-of-the-art approaches.
\end{abstract}

\input{sections/main/intro}
\input{sections/main/related}

\input{sections/main/background}
\input{sections/main/approach}

\input{sections/main/experiments}

\input{sections/main/conclusions}

\clearpage

\bibliography{strings,main}
\bibliographystyle{icml2019}

\end{document}

%% file: math_commands.tex
\usepackage{amsmath,amsfonts,bm}

\def\Figref#1{Figure~\ref{#1}}

\def\Secref#1{Section~\ref{#1}}

\def\eqref#1{equation~\ref{#1}}
\def\Eqref#1{Equation~\ref{#1}}

\def\1{\bm{1}}

\def\vs{{\bm{s}}}

\DeclareMathAlphabet{\mathsfit}{\encodingdefault}{\sfdefault}{m}{sl}
\SetMathAlphabet{\mathsfit}{bold}{\encodingdefault}{\sfdefault}{bx}{n}

%% file: sections/main/intro.tex
\vspace{-20pt}
\section{Introduction}
\label{sec:intro}

Effective communication is a key ability for collaboration.
Indeed, intelligent agents (humans or artificial) in real-world scenarios can
significantly benefit from exchanging information that enables them to coordinate,
strategize, and utilize their combined sensory experiences to act in the physical world.
The ability to communicate has wide-ranging applications for artificial agents --
from multi-player gameplay in simulated (\eg DoTA, StarCraft) or
physical worlds (\eg robot soccer), to self-driving car networks communicating
with each other to achieve safe and swift transport, to teams of robots on
search-and-rescue missions deployed in hostile, fast-evolving environments.

A salient property of human communication is the ability to hold \emph{targeted} interactions.
Rather than the `one-size-fits-all' approach of broadcasting messages to all participating agents,
as has been previously explored~\citep{sukhbaatar_nips16,foerster_nips16,singh_iclr19},
it can be useful to direct certain messages to specific recipients.
This enables a more flexible collaboration strategy in complex environments.
For example, within a team of search-and-rescue robots with a diverse set of roles and goals,
a message for a fire-fighter (\eg ``smoke is coming from the kitchen'') is largely
meaningless for a bomb-defuser.

\begin{table*}[t]
    \vspace{-5pt}
    \setlength\tabcolsep{6pt}
    \centering
    \begin{tabular}{@{}l c c c r@{}}
        &  \footnotesize Decentralized  & \footnotesize Targeted & \footnotesize Multi-Round & \footnotesize Reinforcement  \\[-3pt]
        & \footnotesize Execution & \footnotesize  Communication & \footnotesize Decisions & \footnotesize Learning \\
        \cmidrule(r){1-1} \cmidrule(lr){2-2} \cmidrule(lr){3-3} \cmidrule(lr){4-4} \cmidrule(lr){5-5}
        DIAL~\citep{foerster_nips16}
            & Yes & No & No & Yes (Q-Learning) \\
        CommNet~\citep{sukhbaatar_nips16}
            & Yes & No & Yes & Yes (REINFORCE) \\
        VAIN~\citep{hoshen_nips17}
            & No & Yes & Yes & No (Supervised) \\
        ATOC~\citep{jiang_nips18}
            & Yes & No & No & Yes (Actor-Critic) \\
        IC3Net~\citep{singh_iclr19}
            & Yes & No & Yes & Yes (REINFORCE) \\ \midrule
        TarMAC (this paper) & Yes & Yes & Yes & Yes (Actor-Critic) \\ \bottomrule
    \end{tabular}
    \vspace{5pt}
    \caption{Comparison with previous work on collaborative multi-agent communication with continuous vectors.}
    \vspace{-15pt}
    \label{table:related}
\end{table*}

We develop TarMAC, a Targeted Multi-Agent Communication
architecture for collaborative multi-agent deep reinforcement learning.
Our key insight in TarMAC is to allow each individual agent to \emph{actively select}
which other agents to address messages to. This targeted communication
behavior is operationalized via a simple signature-based soft attention mechanism: along with the message, the sender broadcasts
a key which encodes properties of agents the message is intended for, and is used by receivers to gauge the relevance of the message.
This communication mechanism is learned implicitly, without any attention supervision,
as a result of end-to-end training using task reward.

The inductive bias provided by soft attention in the communication architecture is sufficient to enable agents to 1) communicate agent-goal-specific messages (\eg guide fire-fighter towards fire, bomb-defuser towards
bomb, \etc), 2) be adaptive to variable team sizes (\eg the size of the local neighborhood a self-driving car can communicate with changes as it moves), and 3) be interpretable through predicted attention probabilities that allow for inspection of \emph{which} agent is communicating \emph{what} message and to \emph{whom}.

Our results however show that just using targeted communication is not enough.
Complex real-world tasks might require \emph{large populations of agents} to go through
\emph{multiple rounds of collaborative communication and reasoning}, involving
large amounts of information to be \emph{persistent in memory} and exchanged via
\emph{high-bandwidth communication channels}. To this end, our actor-critic
framework combines centralized training with decentralized execution~\citep{lowe_nips17},
thus enabling scaling to large team sizes. In this context, our inter-agent
communication architecture also supports multiple rounds of targeted interactions at
every time-step, wherein the agents' recurrent policies persist
relevant information in internal states.

While natural language, \ie a finite set of discrete tokens with pre-specified human-conventionalized meanings, may seem like an intuitive protocol for inter-agent communication -- one that enables human-interpretability of interactions -- forcing machines to communicate among themselves in discrete tokens presents additional training challenges. Since our work focuses on machine-only multi-agent teams, we allow agents to communicate via continuous vectors (rather than discrete symbols),
as has been explored in~\cite{sukhbaatar_nips16,singh_iclr19},
and agents have the flexibility to discover and optimize their communication protocol as per task requirements.

We provide extensive empirical evaluation of our approach across
a range of tasks, environments, and team sizes.
\begin{compactitem}
\item We begin by benchmarking TarMAC and its ablation without attention on a
    cooperative navigation task derived from the SHAPES environment~\citep{andreas_cvpr16} in \Secref{sec:shapes}.
    We show that agents learn intuitive attention behavior across task difficulties.
\item Next, we evaluate TarMAC on the traffic junction environment~\citep{sukhbaatar_nips16} in \Secref{sec:tj},
    and show that agents are able to adaptively focus on `active' agents in the
    case of varying team sizes.
\item We then demonstrate its efficacy in $3$D environments with a cooperative first-person point-goal navigation task in House$3$D~\citep{house3d} (\Secref{sec:h3d}).
\item Finally, in \Secref{sec:competitive}, we show that TarMAC can be easily combined with IC3Net~\cite{singh_iclr19},
    thus extending its applicability to mixed and competitive environments, and
    leading to significant improvements in performance and sample complexity.
\end{compactitem}

%% file: sections/main/related.tex
\section{Related Work}
\label{sec:related}

Multi-agent systems fall at the intersection of game theory, distributed systems, and Artificial Intelligence in general~\citep{shoham_08}, and thus have a rich and diverse literature. Our work builds on and is related to prior work in deep multi-agent reinforcement learning, the centralized training and decentralized execution paradigm, and emergent communication protocols.

\textbf{Multi-Agent Reinforcement Learning (MARL).} Within MARL (see \citet{busoniu_08} for a survey), our work
is related to efforts on using recurrent neural networks to approximate agent policies~\citep{hausknecht_aaai15},
stabilizing algorithms for multi-agent training~\citep{lowe_nips17, foerster_aaai18},
and tasks in novel domains \eg coordination and navigation in 3D environments~\citep{peng_arxiv17,openai_18,jaderberg_arxiv18}.

\textbf{Centralized Training \& Decentralized Execution.} Both~\citet{sukhbaatar_nips16} and~\citet{hoshen_nips17}
adopt a centralized framework at both training and test time -- a central controller processes
local observations from all agents and outputs a probability distribution over joint actions.
In this setting, the controller (\eg a fully-connected network) can be viewed as implicitly
encoding communication. \citet{sukhbaatar_nips16} propose an efficient controller architecture
that is invariant to agent permutations by virtue of weight-sharing and averaging (as in~\citet{zaheer_nips17}),
and can, in principle, also be used in a decentralized manner at test time since each
agent just needs its local state vector and the average of incoming messages to take an action.
Meanwhile,~\citet{hoshen_nips17}
proposes to replace averaging by an attentional mechanism to allow targeted interactions between agents.
While closely related to our communication architecture, this work only considers fully-supervised
one-next-step prediction tasks, while we study the full reinforcement learning problem
with tasks requiring planning over long time horizons.

Moreover, a centralized controller quickly becomes intractable in real-world tasks with many agents and high-dimensional observation spaces \eg navigation in House3D~\citep{house3d}.
To address these weaknesses, we adopt the framework of centralized learning but decentralized execution
(following~\citet{foerster_nips16,lowe_nips17}) and further relax it by allowing agents to communicate.
While agents can use extra information during training, at test time, they pick actions solely based on
local observations and communication messages.

\textbf{Emergent Communication Protocols.} Our work is also related to recent work on learning communication
protocols in a completely end-to-end manner with reinforcement learning -- from perceptual input (\eg pixels)
to communication symbols (discrete or continuous) to actions (\eg navigating in an environment).
While~\citep{foerster_nips16,jorge_iclrw16,visdial_rl,kottur_emnlp17,mordatch_arxiv17,lazaridou_iclr17}
constrain agents to communicate with discrete symbols with the explicit goal to study emergence of language,
our work operates in the paradigm of learning a continuous communication protocol in order to solve a downstream
task~\citep{sukhbaatar_nips16,hoshen_nips17,jiang_nips18,singh_iclr19}.
\citet{jiang_nips18,singh_iclr19} also operate in a decentralized execution
setting and use an attentional communication mechanism, but in contrast to our work,
they use attention to decide \emph{when} to communicate, not \emph{who} to communicate with.
In \Secref{sec:competitive}, we discuss how to potentially combine the two approaches.

Table~\ref{table:related} summarizes
the main axes of comparison between our work and previous efforts in this exciting space.

%% file: sections/main/background.tex
\vspace{-10pt}
\section{Technical Background}
\label{sec:background}

\textbf{Decentralized Partially Observable Markov Decision Processes (Dec-POMDPs).}
A Dec-POMDP is a multi-agent extension of a partially observable Markov
decision process \cite{oliehoek_book12}. For $N$ agents, it is defined by a set of
states $S$ describing possible configurations of all agents, a global reward function
$R$, a transition probability function $T$, and for each agent $i \in {1,...,N}$ a
set of allowed actions $A_i$, a set of possible observations $\Omega_i$ and an
observation function $O_i$.
At each time step every agent picks an action $a_i$ based on its local observation
$\omega_i$ following its own stochastic policy $\pi_{\theta_i}(a_i | \omega_i)$.
The system randomly transitions to the next state $s'$ given the current state and
joint action $T(s' | s, a_1,...,a_N)$. The agent team receives a global reward
$r = R(s, a_1,...,a_N)$ while each agent receives a local observation of the new
state $O_i(\omega_i | s')$. Agents aim to maximize the total expected return
$J = \sum_{t=0}^T \gamma^t r_t$ where $\gamma$ is a discount factor and $T$ is
the episode time horizon.

\textbf{Actor-Critic Algorithms.} Policy gradient methods directly adjust the parameters
$\theta$ of the policy in order to maximize the objective
$J(\theta) = \mathbb{E}_{s \sim p_{\pi}, a \sim \pi_{\theta}(s)} \left[ R(s, a) \right]$
by taking steps in the direction of $\nabla J(\theta)$. We can write the gradient
with respect to the policy parameters as the following:
$$\nabla_{\theta} J(\theta) = \mathbb{E}_{s \sim p_{\pi}, a \sim \pi_{\theta}(s)} \left[ \nabla_{\theta }\log \pi_{\theta}(a | s) Q_{\pi} (s, a) \right],$$
where $Q_{\pi} (s, a)$ is the action-value. It is the expected remaining
discounted reward if we take action $a$ in state $s$ and follow policy $\pi$ thereafter.
Actor-Critic algorithms learn an approximation $\hat{Q}(s, a)$ of the unknown true
action-value function by e.g. temporal-difference learning~\citep{sutton_book98}.
This $\hat{Q}(s, a)$ is the Critic and $\pi_{\theta}$ is the Actor.

\textbf{Multi-Agent Actor-Critic.} \citet{lowe_nips17} propose a multi-agent Actor-Critic
algorithm adapted to centralized learning and decentralized execution wherein each
agent learns its own policy $\pi_{\theta_i}(a_i | \omega_i)$ conditioned on local
observation $\omega_i$ using a central Critic that estimates the joint
action-value $\hat{Q}(s, a_1, ..., a_N)$ conditioned on all actions.

%% file: sections/main/approach.tex
\section{TarMAC: Targeted Multi-Agent Communication}
\label{sec:approach}

\begin{figure*}[t]
    \centering
    \includegraphics[width=\textwidth]{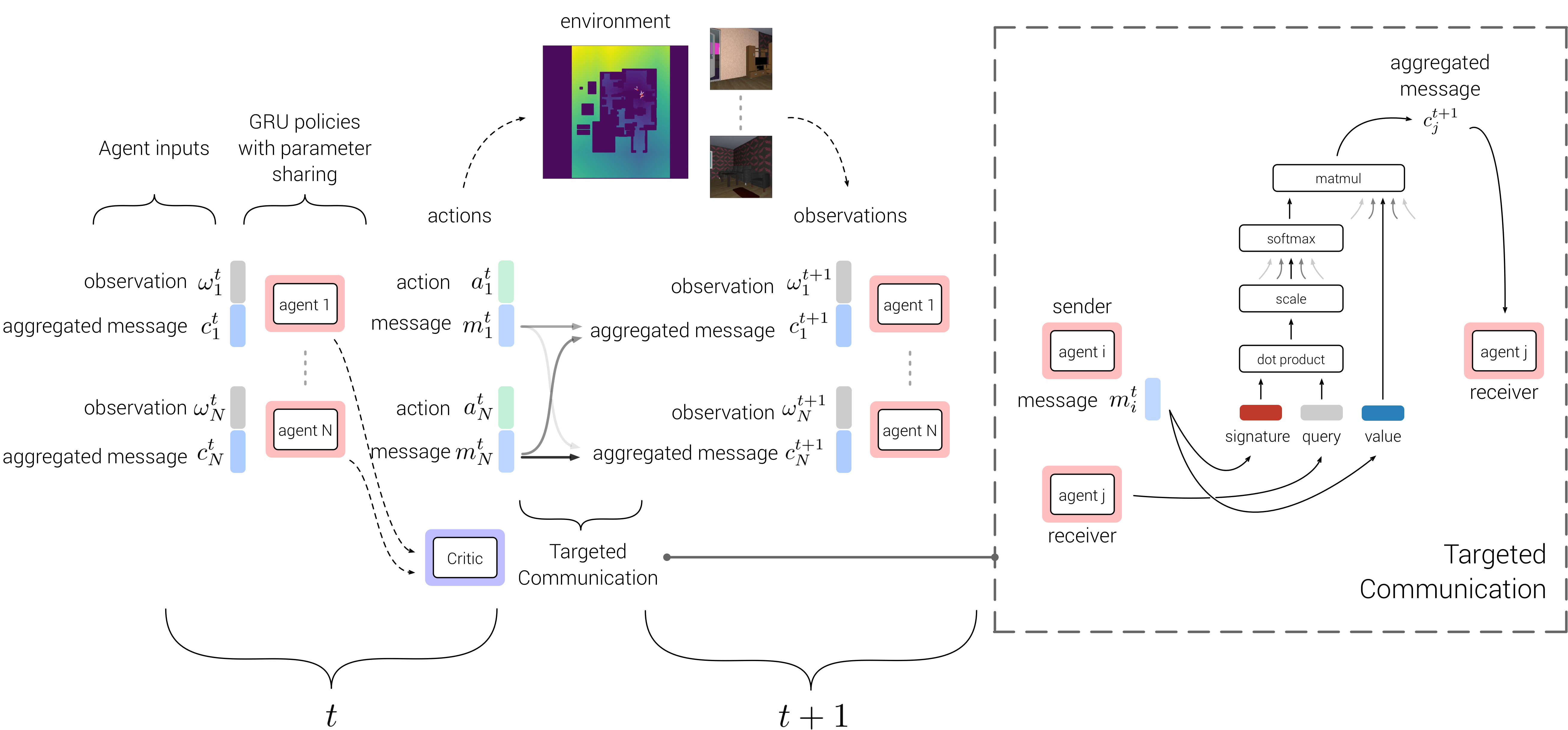}
    \vspace{2pt}
    \caption{Overview of our multi-agent architecture with targeted communication.
    Left: At every timestep, each agent policy gets a local observation $\omega_i^t$
    and aggregated message $c_i^t$ as input, and predicts an environment action $a_i^t$
    and a targeted communication message $m_i^t$. Right: Targeted communication
    between agents is implemented as a signature-based soft attention mechanism.
    Each agent broadcasts a message $m_i^t$ consisting of a signature $k_i^t$,
    which can be used to encode agent-specific information and
    a value $v_i^t$, which contains the actual message. At the next timestep, each receiving agent
    gets as input a convex combination of message values, where the
    attention weights are obtained by a dot product between sender's signature $k_i^t$
    and a query vector $q_j^{t+1}$ predicted from the receiver's hidden state.}
    \vspace{-15pt}
\end{figure*}

We now describe our multi-agent communication architecture in detail.
Recall that we have $N$ agents with policies $\{\pi_1, ..., \pi_N\}$, respectively
parameterized by $\{\theta_1, ..., \theta_N\}$, jointly performing a cooperative task.
At every timestep $t$, the $i$th agent for all $i \in \{1, ..., N\}$ sees a local
observation $\omega_i^t$, and must select a discrete environment action $a_i^t \sim \pi_{\theta_i}$
and send a continuous communication message $m_i^t$, received by other agents at the next timestep,
in order to maximize global reward $r_t \sim R$. Since no agent has access to the
underlying complete state of the environment $s_t$, there is incentive in communicating
with each other and being mutually helpful to do better as a team.

\textbf{Policies and Decentralized Execution.} Each agent is essentially modeled as a
Dec-POMDP augmented with communication.  Each agent's policy $\pi_{\theta_i}$
is implemented as a $1$-layer Gated Recurrent Unit~\citep{cho_emnlp14}.
At every timestep, the local observation $\omega_i^t$ and a vector $c_i^t$
aggregating messages sent by all agents at the previous timestep (described in
more detail below) are used to update the hidden state $h_i^t$ of the GRU,
which encodes the entire message-action-observation history up to time $t$.
From this internal state representation, the agent's policy
$\pi_{\theta_i} \left(a_i^t \, | \, h_i^t \right)$ predicts a categorical
distribution over the space of actions, and another output head produces an
outgoing message vector $m_i^t$. Note that for our experiments,
agents are symmetric and policy parameters are shared across agents,
\ie $\theta_1 = ... = \theta_N$.
This considerably speeds up learning.

\textbf{Centralized Critic.} Following prior work~\citep{lowe_nips17,foerster_aaai18},
we operate under the centralized learning and decentralized execution
paradigm wherein during training, a centralized Critic
guides the optimization of individual agent policies. The Critic takes as
input predicted actions $\{ a_1^t, ..., a_N^t \}$ and internal state
representations $\{h_1^t, ..., h_N^t \}$ from all agents to estimate the
joint action-value $\hat Q_t$ at every timestep. The centralized Critic
is learned by temporal difference~\citep{sutton_book98} and the gradient
of the expected return $J(\theta_i) = \mathbb{E}[R]$ with respect to
policy parameters is approximated by:

\resizebox{1. \columnwidth}{!}
{
$
    \nabla_{\theta_{i}} J(\theta_i) =
        \mathbb{E}
            \left[ \nabla_{\theta_{i}} \log \pi_{\theta_{i}} (a_i^t | h_i^t)
                \,\, \hat{Q}_{t} (h_1^t, ..., h_N^t, a_t^1, ..., a_t^N) \right].
$
}

Note that compared to an individual Critic $\hat{Q}_i(h_i^t, a_i^t)$ per agent,
having a centralized Critic leads to considerably lower variance
in policy gradient estimates since it takes into account actions from all agents.
At test time, the Critic is not needed and policy execution is
fully decentralized.

\textbf{Targeted, Multi-Round Communication.}
Establishing complex collaboration strategies requires targeted communication
\ie the ability to address specific messages to specific agents, as well as
multi-round communication \ie multiple rounds of back-and-forth interactions between
agents. We use a signature-based soft-attention mechanism in our communication
structure to enable targeting. Each message $m_i^t$ consists of $2$ parts:
a signature $k_i^t \in \mathbb{R}^{d_k}$ to encode properties of intended recipients
and a value $v_i^t \in \mathbb{R}^{d_v}$:
\begin{equation}\label{eq:message}
m_i^t = [\,\,\,\overbracket{\,\,\,\,\,\, k_i^t \,\,\,\,\,\,}^{\textnormal{signature}} \quad
    \underbracket{\,\,\, v_i^t \,\,\,}_{\textnormal{value}}\,\,\,] \;.
\end{equation}
At the receiving end, each agent (indexed by $j$) predicts a query
vector $q_j^{t+1} \in \mathbb{R}^{d_k}$ from its hidden state $h_j^{t+1}$,
which is used to compute a dot product with signatures of all $N$ messages.
This is scaled by $1 / \sqrt{d_k}$ followed by a softmax to
obtain attention weights $\alpha_{ji}$ for each incoming message:
\begin{equation}\label{eq:alpha}
\mathbf{\alpha}_{j} = \textnormal{softmax}\left[
\frac{{q_j^{t+1}}^T {k_1^t}}{\sqrt{d_k}} \,\, ... \,\,
    \underbracket{\frac{ {{q_j^{t+1}}^T {k_i^t}}}{\sqrt{d_k}}}_{\alpha_{ji}} \,\, ... \,\, \frac{{q_j^{t+1}}^T {k_N^t}}{\sqrt{d_k}} \right] \quad \quad \quad
\end{equation}

\begin{figure*}[t!]
    \centering
    \includegraphics[width=\textwidth]{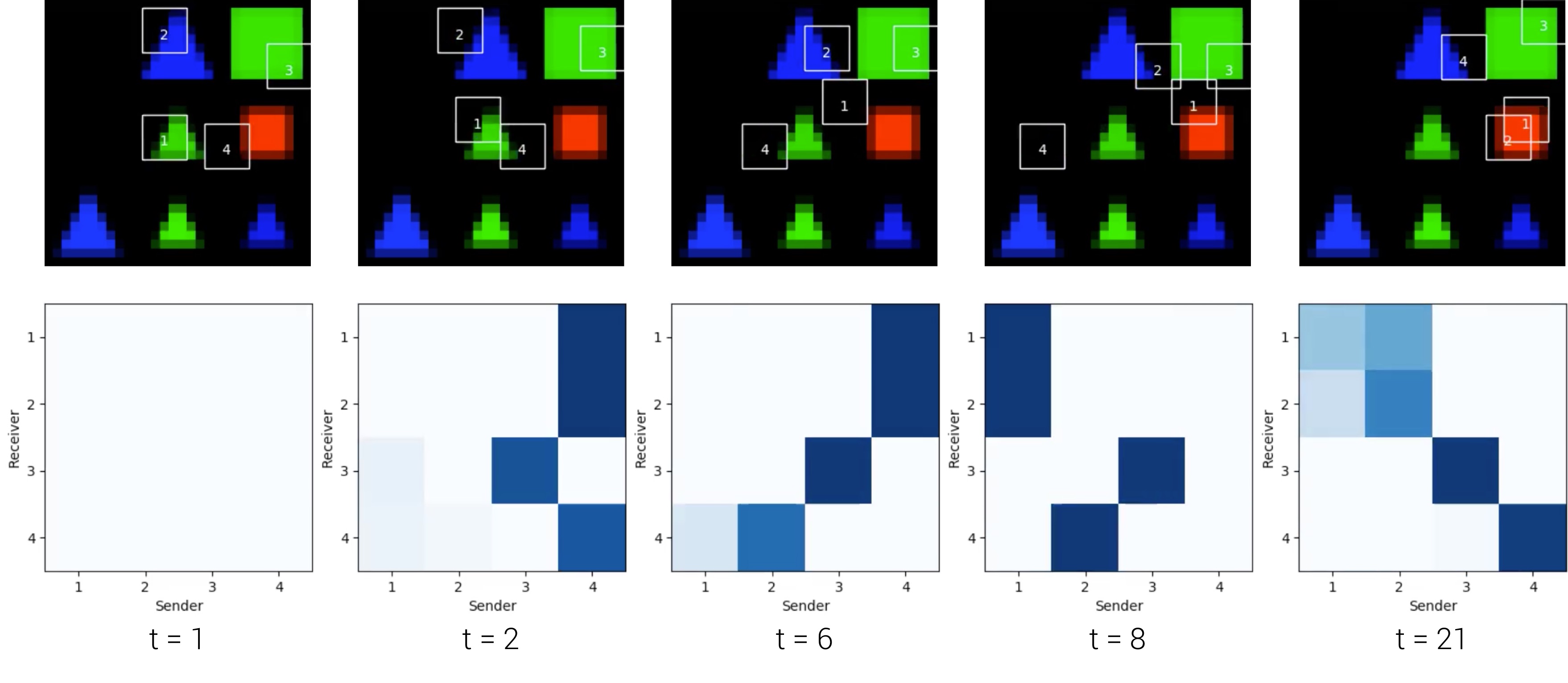}
    \vspace{-15pt}
    \caption{Visualizations of learned targeted communication in SHAPES. Figure best viewed in color.
    $4$ agents have to find $[$\texttt{red}, \texttt{red}, \texttt{green}, \texttt{blue}$]$ respectively.
    $t=1$: inital spawn locations;
    $t=2$: $4$ was on \texttt{red} at $t=1$ so $1$ and $2$ attend to messages from $4$ since they have to find \texttt{red}. $3$ has found its goal (\texttt{green}) and is self-attending;
    $t=6$: $4$ attends to messages from $2$ as $2$ is on $4$'s target -- \texttt{blue};
    $t=8$: $1$ finds \texttt{red}, so $1$ and $2$ shift attention to $1$;
    $t=21$: all agents are at their respective goal locations and primarily self-attending.}
    \label{fig:shapes_3x3}
    \vspace{-15pt}
\end{figure*}
used to compute
$c_j^{t+1}$, the input message for agent $j$ at $t+1$:
\begin{equation}\label{eq:attended}
c_j^{t+1} = \sum_{i=1}^N \alpha_{ji} v_i^t.
\end{equation}
Intuitively, attention weights are high when both sender and receiver
predict similar signature and query vectors respectively.
Note that \Eqref{eq:alpha} also includes $\alpha_{ii}$ corresponding
to the ability to \emph{self-attend}~\citep{vaswani_nips17}, which we empirically found
to improve performance, especially in situations when an agent has found
the goal in a coordinated navigation task and all it is required to
do is stay at the goal, so others benefit from attending to this agent's
message but return communication is not necessary.
Note that the targeting mechanism in our formulation is implicit
\ie agents implicitly encode properties without addressing specific recipients.
For example, in a self-driving car network, a particular message may be for
``cars travelling on the west to east road" (implicitly encoding properties)
as opposed to specifically for ``car 2'' (explicit addressing).

For multi-round communication, aggregated message vector $c_j^{t+1}$ and
internal state $h_j^t$ are first used to predict the next internal state
${h'}_j^t$ taking into account the first round:
\begin{equation}\label{eq:nextstate}
{h'}_j^t = \tanh\left( W_{h\rightarrow h'} [ \,\, c_j^{t+1} \,\, \Vert \,\, h_j^t \,\, ]  \right).
\end{equation}
Next, the updated hidden state ${h'}_j^t$ is used to predict signature, query, value
followed by repeating Equations~\refeq{eq:message}-\refeq{eq:nextstate} for
multiple rounds until we get a final aggregated message vector $c_j^{t+1}$
to be used as input at the next timestep.
Number of rounds of communication is treated as a hyperparameter.

Our entire communication architecture is differentiable, and
message vectors are learnt through backpropagation.

%% file: sections/main/experiments.tex
\section{Experiments}
\label{sec:experiments}

We evaluate TarMAC on a variety of tasks and environments.
All our models were trained with a batched synchronous version of the multi-agent
Actor-Critic described above, using RMSProp with a learning rate of
$7\times 10^{-4}$ and $\alpha = 0.99$, batch size $16$, discount factor
$\gamma = 0.99$ and entropy regularization coefficient $0.01$ for agent policies.
All our agent policies are instantiated from the same set of shared parameters; \ie $\theta_1 = ... = \theta_N$.
Each agent's GRU hidden state is $128$-d, message signature/query is $16$-d, and message value is $32$-d (unless specified otherwise).
All results are averaged over $5$ independent seeds (unless noted otherwise),
and error bars show standard error of means.

\vspace{-5pt}
\subsection{SHAPES}
\label{sec:shapes}

The SHAPES dataset was introduced by~\citet{andreas_cvpr16}\footnote{\href{https://github.com/jacobandreas/nmn2/tree/shapes}{\texttt{github.com/jacobandreas/nmn2/tree/shapes}}}, and originally created
for testing compositional visual reasoning for the task of visual question answering.
It consists of synthetic images of $2$D colored shapes
arranged in a grid ($3 \times 3$ cells in the original dataset) along with corresponding
question-answer pairs.
There are $3$ shapes (circle, square, triangle), $3$ colors (red, green, blue),
and $2$ sizes (small, big) in total (see~\Figref{fig:shapes_3x3}).

We convert each image from SHAPES
into an active environment where agents can now be spawned at different regions of
the image, observe a $5 \times 5$ local patch around them and their coordinates, and take actions
to move around -- $\{$\texttt{up}, \texttt{down}, \texttt{left}, \texttt{right}, \texttt{stay}$\}$. Each agent is tasked with
navigating to a specified goal state in the environment within a max no.~of steps -- $\{$`red', `blue square',
`small green circle', \etc$\}$ -- and the reward for each agent at every timestep
is based on team performance \ie $r_t = \frac{\textnormal{\# agents on goal}}{\textnormal{\# agents}}$.
Having a symmetric, team-based reward incentivizes agents to cooperate
in finding each agent's goal.

\begin{table*}[h!]
\centering
\begin{tabular}{lrrr}
   & \multicolumn{1}{c}{{\scriptsize $30 \times 30$, $4$ agents, find$[$\texttt{red}$]$}} & \multicolumn{1}{c}{{\scriptsize $50 \times 50$, $4$ agents, find$[$\texttt{red}$]$}} & \multicolumn{1}{c}{{\scriptsize $50 \times 50$, $4$ agents, find$[$\texttt{red,red,green,blue}$]$}}    \\ \midrule
    {\small No communication}
        & {\small $95.3$}{\scriptsize $\pm 2.8\%$} & {\small $83.6$}{\scriptsize $\pm 3.3\%$} & {\small $69.1$}{\scriptsize $\pm 4.6\%$}  \\[0.05in]
    {\small No attention}
        & {\small $\mathbf{99.7}$}{\scriptsize $\pm 0.8\%$} & {\small $\mathbf{89.5}$}{\scriptsize $\pm 1.4\%$} & {\small $82.4$}{\scriptsize $\pm 2.1\%$} \\[0.05in]
    {\small TarMAC}
        & {\small $\mathbf{99.8}$}{\scriptsize $\pm 0.9\%$} & {\small $\mathbf{89.5}$}{\scriptsize $\pm 1.7\%$} & {\small $\mathbf{85.8}$}{\scriptsize $\pm 2.5\%$}  \\ \bottomrule
\end{tabular} \\[0.05in]
\caption{Success rates on $3$ different settings of cooperative navigation
    in the SHAPES environment.}
\label{table:shapesnumbers}
\vspace{-15pt}
\end{table*}

\textbf{How does targeting work?}
Recall that each agent predicts a signature and value vector as part of the message
it sends, and a query vector to attend to incoming messages.
The communication is targeted because the attention probabilities are a function
of both the sender's signature and receiver's query vectors. So it is not just
the receiver deciding how much of each message to listen to.
The sender also sends out signatures that affects how much of each
message is sent to each receiver.
The sender's signature could encode parts of its observation most relevant to other
agents' goals (\eg it would be futile to convey coordinates in the signature),
and the message value could contain the agent's own location. For example, in ~\Figref{fig:shapes_3x3},
at $t=6$, we see that when agent 2 passes by blue, agent 4 starts attending to agent 2. Here,
agent 2's signature likely encodes the color it observes (which is blue), and agent 4's query
encodes its goal (which is also blue) leading to high attention probability.
Agent 2's message value encodes coordinates agent 4 has to navigate to,
which it ends up reaching by $t=21$.

SHAPES serves as a flexible testbed for carefully controlling and analyzing
the effect of changing the size of the environment, no. of agents, goal configurations, \etc.
\Figref{fig:shapes_3x3} visualizes learned protocols, and \tableref{table:shapesnumbers}
reports quantitative evaluation for three different configurations --
1) 4 agents, all tasked with finding \texttt{red} in $30 \times 30$ images,
2) 4 agents, all tasked with finding \texttt{red} in $50 \times 50$ images,
3) 4 agents, tasked with finding $[$\texttt{red,red,green,blue}$]$ respectively in $50 \times 50$ images.
We compare TarMAC against two baselines -- 1) without communication, and 2) with communication
but where broadcasted messages are averaged instead of attention-weighted, so
all agents receive the same message vector, similar to~\citet{sukhbaatar_nips16}.
Benefits of communication and attention increase with task complexity
({\scriptsize $30 \times 30 \rightarrow 50 \times 50$ \& find$[$\texttt{red}$]$ $\rightarrow$ find$[$\texttt{red,red,green,blue}$]$}).

\vspace{-5pt}
\subsection{Traffic Junction}
\label{sec:tj}

\textbf{Environment and Task.} The simulated traffic junction environments from~\cite{sukhbaatar_nips16} consist of
cars moving along pre-assigned, potentially intersecting routes on one or more road junctions. The total number of cars is fixed at $N_\textnormal{max}$,
and at every timestep new cars get added to the environment with probability $p_\textnormal{arrive}$. Once a car completes its route, it becomes available to be sampled and added
back to the environment with a different route assignment.
Each car has a limited visibility of a $3 \times 3$ region around it, but is free
to communicate with all other cars. The action space for each car
at every timestep is \texttt{gas} and \texttt{brake}, and the reward consists of a linear time penalty
$-0.01\tau$, where $\tau$ is the number of timesteps since car has been active,
and a collision penalty $r_\textnormal{collision} = -10$.

\begin{table}[h]
\centering
    \resizebox{\columnwidth}{!}{
    \begin{tabular}{lrr}
    \toprule
       & \multicolumn{1}{c}{Easy} & \multicolumn{1}{c}{Hard}    \\ \midrule
        No communication
            & $84.9${\scriptsize $\pm 4.3\%$}  & $74.1${\scriptsize $\pm 3.9\%$}  \\[0.05in]
        CommNet~\citep{sukhbaatar_nips16}
            & $99.7${\scriptsize $\pm 0.1\%$}  & $78.9${\scriptsize $\pm 3.4\%$}  \\[0.05in]
        TarMAC $1$-round
            & $\mathbf{99.9}${\scriptsize $\pm 0.1\%$}   & $84.6${\scriptsize $\pm 3.2\%$}  \\[0.05in]
        TarMAC $2$-round
            & $\mathbf{99.9}${\scriptsize $\pm 0.1\%$} & $\mathbf{97.1}${\scriptsize $\pm 1.6\%$}  \\ \bottomrule
    \end{tabular}} \\[0.05in]
    \caption{Success rates on traffic junction. Our targeted
    $2$-round communication architecture gets a success rate of $97.1${\scriptsize $\pm 1.6\%$}
    on the `hard' variant, significantly outperforming~\citet{sukhbaatar_nips16}.
    Note that $1$- and $2$-round refer to the number of rounds of communication between actions (\Eqref{eq:nextstate}).}
    \label{table:traffic}
    \vspace{-5pt}
\end{table}

\begin{figure*}[t]
    \begin{minipage}[b]{0.28\textwidth}
        \begin{subfigure}{\textwidth}
            \centering
            \includegraphics[width=0.9\textwidth]{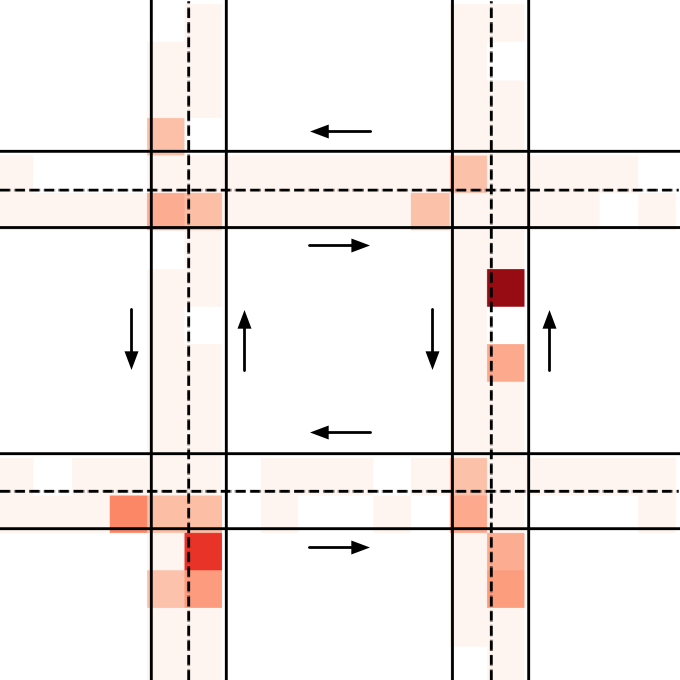}
            \vspace{5pt}
            \caption{Brake probabilities at different locations on the \texttt{hard} traffic
                junction environment. Cars tend to brake close to or right before entering junctions.}
            \label{fig:brake}
        \end{subfigure}
    \end{minipage}\hfill
    \begin{minipage}[b]{0.27\textwidth}
        \begin{subfigure}{\textwidth}
            \centering
            \includegraphics[width=0.9\textwidth]{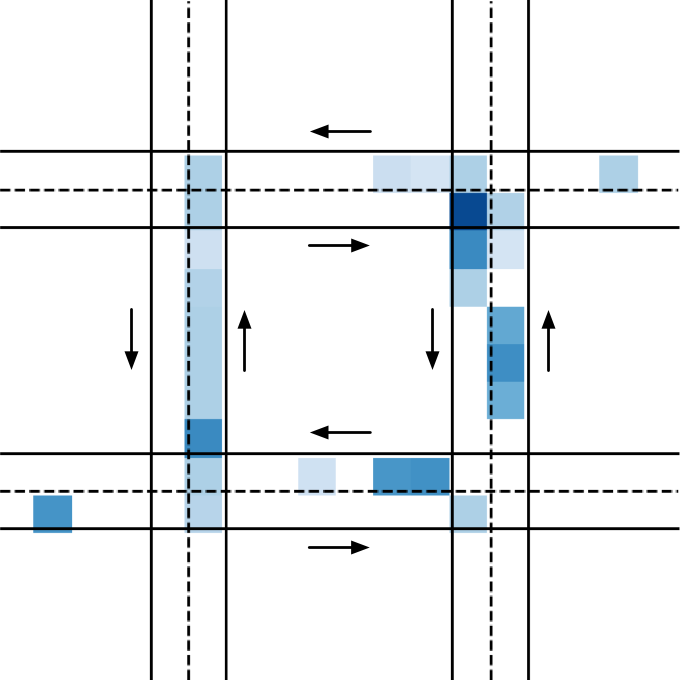}
            \vspace{5pt}
            \caption{Attention probabilities at different locations.
            Cars are most attended to in the `internal grid' -- right after the $1$st
            junction and before the $2$nd.}
            \label{fig:attention}
        \end{subfigure}
    \end{minipage}\hfill
    \begin{subfigure}{0.41\textwidth}
        \centering
        \vspace{15pt}
        \includegraphics[width=0.9\textwidth]{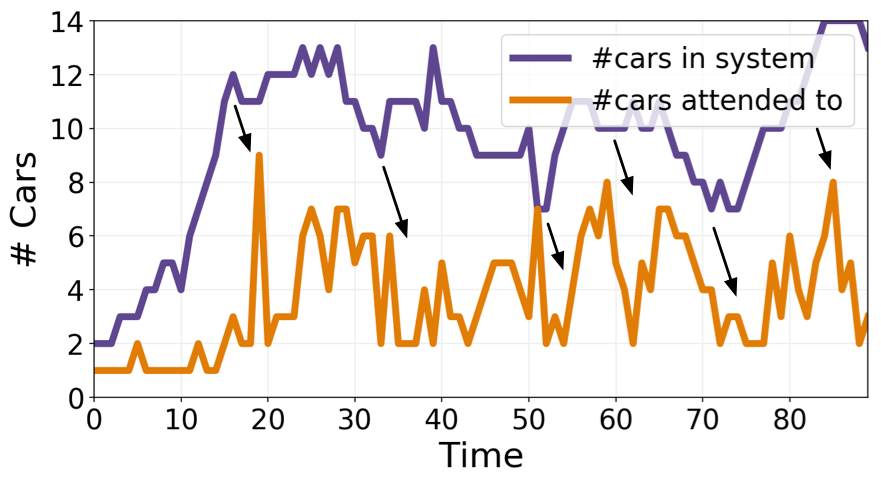}
        \vspace{10pt}
        \caption{No.~of cars being attended to
        1) is positively correlated with total cars,
        indicating that TarMAC is adaptive to dynamic team sizes,
        and 2) is slightly right-shifted, since it takes few steps
        of communication to adapt.}
        \label{fig:carcount}
    \end{subfigure}
    \vspace{5pt}
    \caption{Interpretation of model predictions from TarMAC in the traffic junction environment.}
    \label{fig:traffic}
    \vspace{-10pt}
\end{figure*}

\textbf{Quantitative Results.} We compare our approach with CommNet~\citep{sukhbaatar_nips16} on the \texttt{easy}
and \texttt{hard} difficulties of the traffic junction environment. The \texttt{easy} task has one junction of two one-way roads on a $7 \times 7$ grid with $N_\textnormal{max}=5$ and
$p_\textnormal{arrive}=0.30$, while the \texttt{hard} task has four connected
junctions of two-way roads on a $18 \times 18$ grid with $N_\textnormal{max}=20$ and
$p_\textnormal{arrive}=0.05$.
See \Figref{fig:brake}, \ref{fig:attention} for an example
of the four two-way junctions in the \texttt{hard} task. As shown in \tableref{table:traffic},
a no communication baseline has success rates of $84.9${\scriptsize $\pm 4.3\%$}
and $74.1${\scriptsize $\pm 3.9\%$} on \texttt{easy} and \texttt{hard}
respectively. On \texttt{easy}, both CommNet and TarMAC get close to $100\%$.
On \texttt{hard}, TarMAC with $1$-round significantly outperforms
CommNet with a success rate of $84.6${\scriptsize $\pm 3.2\%$}, while $2$-round
further improves on this at $97.1${\scriptsize $\pm 1.6\%$},
which is an $\sim$$18\%$ absolute improvement over CommNet.
We did not see gains going beyond $2$ rounds in this environment.

\begin{figure}[h]
    \centering
    \includegraphics[width=\columnwidth]{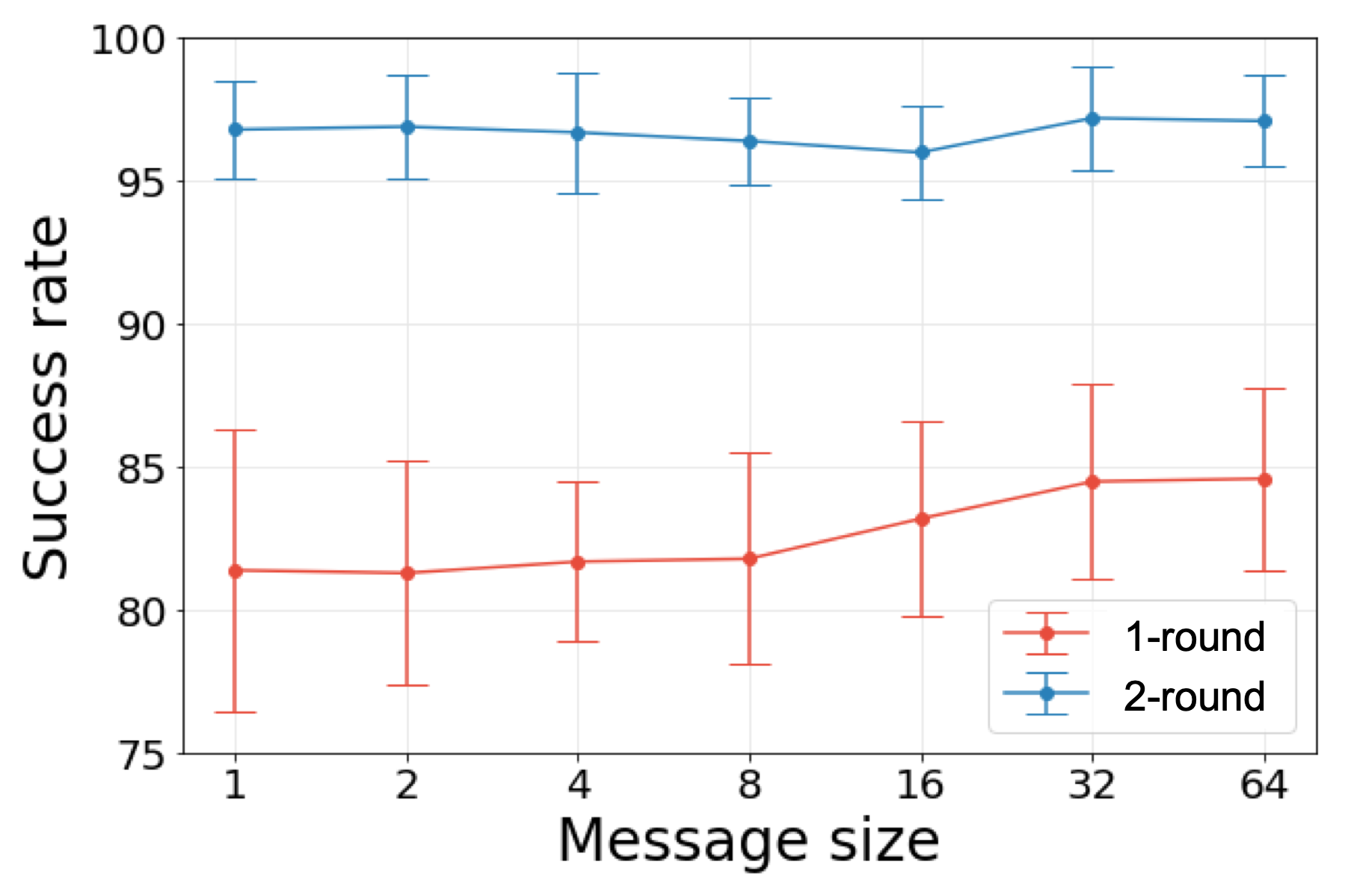}
    \caption{Success rates for $1$ \vs $2$-round \vs message
    size on \texttt{hard}. Performance does not
    decrease significantly even when the message vector is a single scalar, and
    $2$-round communication before taking an
    action leads to significant improvements over $1$-round.}
    \label{fig:msg_sz_rounds}
\end{figure}

\textbf{Message size \vs multi-round communication.}
We study performance of TarMAC with varying message value size and number of rounds
of communication on the \texttt{hard} variant of the traffic junction task. As can be seen in~\Figref{fig:msg_sz_rounds},
multiple rounds of communication leads to significantly higher performance than
simply increasing message size, demonstrating the advantage of multi-round communication.
In fact, decreasing message size to a single scalar performs almost as well as $64$-d,
perhaps because even a single real number can be sufficiently partitioned to cover the space of meanings/messages that need to be conveyed.

\begin{figure*}[h!]
    \centering
    \includegraphics[width=\textwidth]{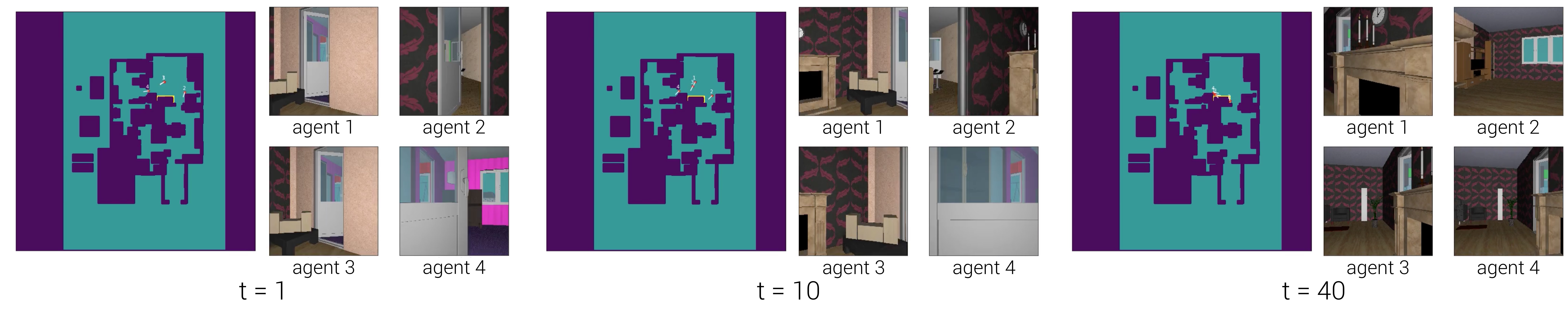}
    \caption{Agents navigating to the \texttt{fireplace} in House3D (marked in yellow).
    Note in particular that agent $4$ is spawned facing away from it.
    It communicates with others, turns to face the \texttt{fireplace}, and moves towards it.}
    \label{fig:house3dnav}
    \vspace{-15pt}
\end{figure*}

\vspace{5pt}
\textbf{Model Interpretation.} Interpreting the learned policies of TarMAC,
\Figref{fig:brake} shows braking probabilities at different locations --
cars tend to brake close to or right before entering traffic junctions, which is reasonable
since junctions have the highest chances for collisions.
Turning our attention to attention probabilities (\Figref{fig:attention}),
we can see that cars are most-attended to when in the `internal grid' -- right after
crossing the $1$st junction and before hitting the $2$nd junction. These attention
probabilities are intuitive -- cars learn to attend to specific sensitive locations
with the most relevant local observations to avoid collisions.
Finally,~\Figref{fig:carcount}
compares total number of cars in the environment
\vs number of cars
being attended to with probability $> 0.1$ at any time. Interestingly, these are (loosely)
positively correlated, with Spearman's $\sigma=0.49$, which shows that TarMAC
is able to adapt to variable number of agents. Crucially, agents learn this dynamic targeting behavior purely from task rewards with
no hand-coding! Note that the right shift between the two curves is
expected, as it takes a few timesteps of communication for team size changes to propagate.
At a relative time shift of $3$, the Spearman's rank correlation between the two curves goes up to $0.53$.

\vspace{-5pt}
\subsection{House3D}
\label{sec:h3d}

Next, we benchmark TarMAC on a cooperative point-goal
navigation task in House3D~\citep{house3d}. House3D provides a rich and diverse
set of publicly-available\footnote{\href{https://github.com/facebookresearch/house3d}{\texttt{github.com/facebookresearch/house3d}}}
$3$D indoor environments, wherein agents do not have access to the top-down map
and must navigate purely from first-person vision. Similar to SHAPES, the agents are tasked
with finding a specified goal (such as `fireplace') within a max no.~of steps,
spawned at random locations in the environment and allowed to communicate and move around.
Each agent gets a shaped reward based on progress towards the specified target.
An episode is successful if all agents end within $0.5$m of the target object
in $500$ navigation steps.

\tableref{table:house3d} shows success rates on a \texttt{find[fireplace]} task
in House3D. A no-communication navigation policy trained with the same
reward structure gets a success rate of $62.1${\scriptsize $\pm 5.3\%$}.
Mean-pooled communication (no attention) performs slightly better with a
success rate of $64.3${\scriptsize $\pm 2.3\%$}, and TarMAC
achieves the best success rate at $68.9${\scriptsize $\pm 1.1\%$}.
TarMAC agents take $82.5$ steps to reach the target on average vs. $101.3$
for no attention vs. $186.5$ for no communication.
\Figref{fig:house3dnav} visualizes a predicted navigation trajectory of $4$ agents.
Note that the communication vectors are significantly more compact ($32$-d)
than the high-dimensional observation space ($224 \times 224$ image),
making our approach particularly attractive for scaling to large agent teams.

\vspace{10pt}
\begin{table}[h]
\centering
\begin{tabular}{lrr}
\toprule
    & \multicolumn{1}{c}{Success rate} & \multicolumn{1}{c}{Avg. \# steps}    \\ \midrule
    No communication
        & $62.1${\scriptsize $\pm 5.3\%$}
        & $186.5$  \\[0.05in]
    No attention
        & $64.3${\scriptsize $\pm 2.3\%$}
        & $101.3$ \\[0.05in]
    TarMAC
        & $\mathbf{68.9}${\scriptsize $\pm 1.1\%$}
        & $\mathbf{82.5}$  \\ \bottomrule
\end{tabular} \\[0.09in]
\caption{$4$-agent \texttt{find[fireplace]} navigation task in House3D.}
\label{table:house3d}
\end{table}

Note that House3D is a challenging testbed for multi-agent reinforcement
learning. To get to $\sim$$100\%$ accuracy, agents have to deal with high-dimensional
visual observations, be able to navigate long action sequences (up to $\sim$$500$ steps),
and avoid getting stuck against objects, doors, and walls.

\vspace{-5pt}
\subsection{Mixed and Competitive Environments}
\label{sec:competitive}

\begin{figure*}[t]
    \begin{minipage}[b]{0.45\textwidth}
        \centering
        \begin{subfigure}{1.0\textwidth}
            \centering
            \includegraphics[width=0.9\textwidth]{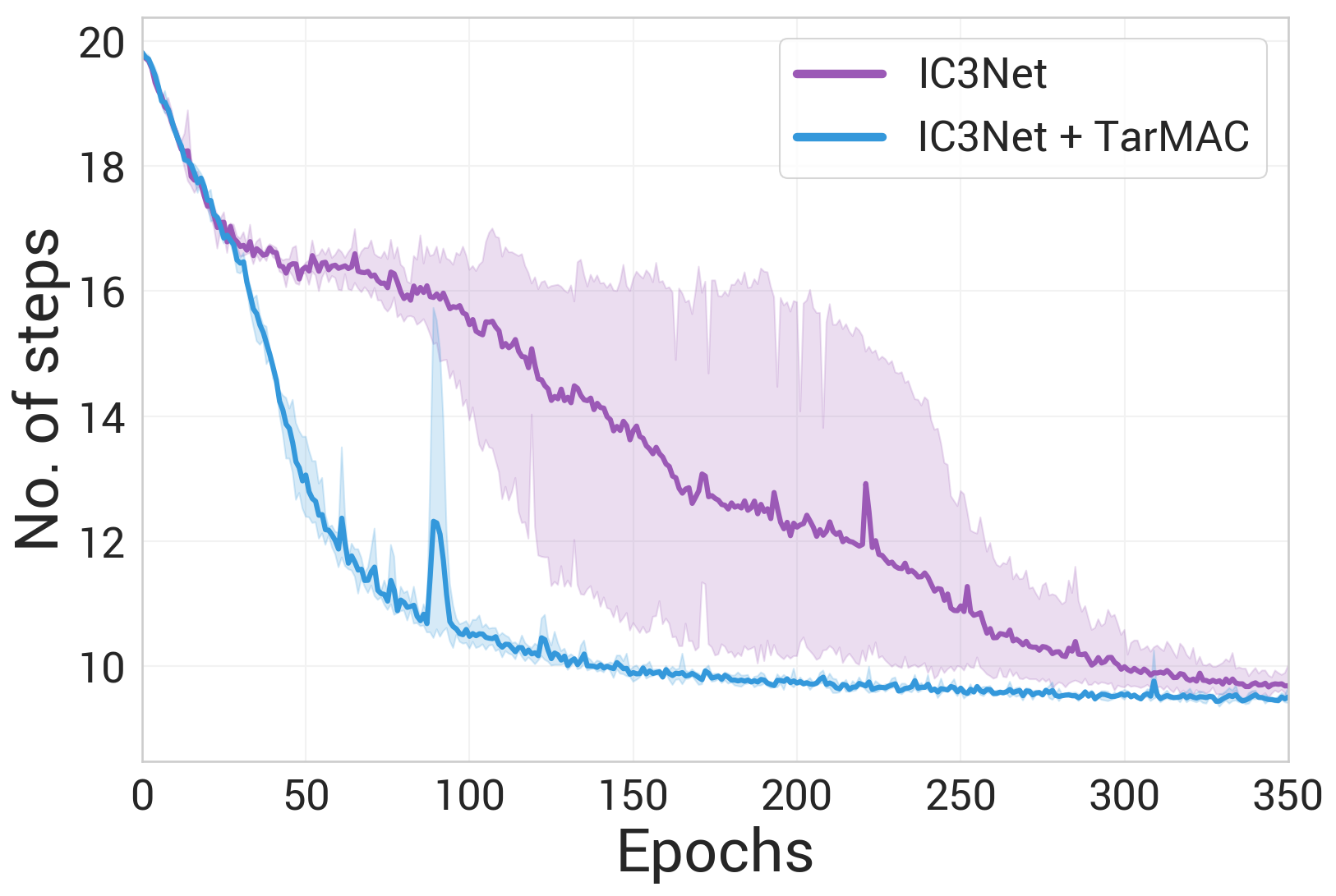}
            \vspace{5pt}
            \caption{$3$ agents, $5\times5$ grid,
                vision=$0$, max steps=$20$}
            \label{fig:pp_easy}
            \vspace{10pt}
        \end{subfigure}
    \end{minipage}\hfill
    \begin{minipage}[b]{0.45\textwidth}
        \centering
        \begin{subfigure}{\textwidth}
            \centering
            \includegraphics[width=0.9\textwidth]{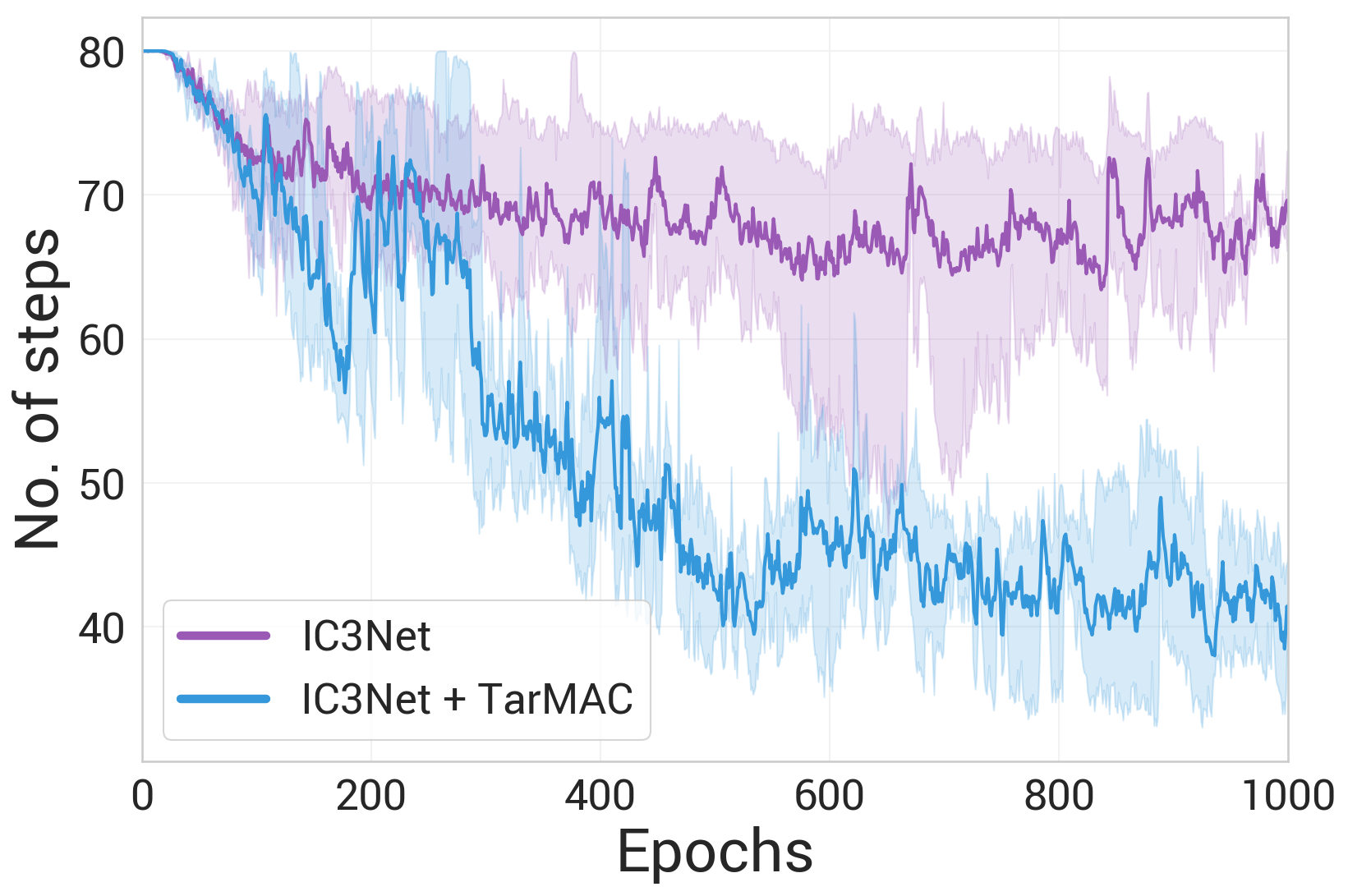}
            \vspace{5pt}
            \caption{$10$ agents, $20\times20$ grid,
                vision=$1$, max steps=$80$}
            \label{fig:pp_hard}
            \vspace{10pt}
        \end{subfigure}
    \end{minipage}\hfill
    \vspace{-5pt}
    \caption{Average no.~of steps to complete an episode (lower is better)
        during training in the Predator-Prey mixed environment.
        IC3Net + TarMAC converges much faster than IC3Net,
        demonstrating that attentional communication helps.
        Shaded region shows $95\%$ CI.}
    \label{fig:pp}
    \vspace{-10pt}
\end{figure*}

\begin{table*}[t]
\centering
\begin{tabular}{lccc}
    & \multicolumn{1}{c}{{\scriptsize $3$ agents, $5 \times 5$,}}
        & \multicolumn{1}{c}{{\scriptsize $5$ agents, $10 \times 10$,}}
        & \multicolumn{1}{c}{{\scriptsize $10$ agents, $20 \times 20$,}} \\[-0.05in]
    & \multicolumn{1}{c}{{\scriptsize vision=$0$, max steps=$20$}}
        & \multicolumn{1}{c}{{\scriptsize vision=$1$, max steps=$40$}}
        & \multicolumn{1}{c}{{\scriptsize vision=$1$, max steps=$80$}}    \\ \midrule
    {CommNet~\cite{sukhbaatar_nips16}}
        & {$9.1$}{\scriptsize $\pm 0.1$} & {$13.1$}{\scriptsize $\pm 0.01$} & {$76.5$}{\scriptsize $\pm 1.3$}  \\ [0.05in]
    {IC3Net~\cite{singh_iclr19}}
        & {$8.9$}{\scriptsize $\pm 0.02$} & {$13.0$}{\scriptsize $\pm 0.02$} & {$52.4$}{\scriptsize $\pm 3.4$} \\ [0.05in]
    {IC3Net $+$ TarMAC}
        & {$8.31$}{\scriptsize $\pm 0.06$} & {$\mathbf{12.74}$}{\scriptsize $\pm 0.08$} & {$41.67$}{\scriptsize $\pm 5.82$} \\ [0.05in]
    {IC3Net $+$ TarMAC (2-round)}
        & {$\mathbf{7.24}$}{\scriptsize $\pm 0.08$} & -- & {$\mathbf{35.57}$}{\scriptsize $\pm 3.96$} \\ \bottomrule
\end{tabular} \\[0.1in]
\caption{Average number of steps taken to complete an episode (lower is better)
        at convergence in the Predator-Prey mixed environment.}
\label{table:ppnumbers}
\vspace{-15pt}
\end{table*}

Finally, we look at how to extend TarMAC to
mixed and competitive scenarios. Communication via sender-receiver soft
attention in TarMAC is poorly suited for competitive scenarios,
since there is always ``leakage'' of the agent's state as a message to other
agents via a low but non-zero attention probability, thus compromising its
strategy and chances of success. Instead, an agent should first be able to independently decide if it wants to
communicate at all, and then direct its message to specific recipients if it does.

The recently proposed IC3Net architecture by~\citet{singh_iclr19} addresses
the former -- learning when to communicate. At every timestep, each agent in IC3Net
predicts a hard gating action to decide if it wants to communicate.
At the receiving end, messages from agents who decide to communicate are
averaged to be the next input message.
Replacing this message averaging with our sender-receiver soft attention,
while keeping the rest of the architecture and training details the same as
IC3Net,
should provide an inductive bias for more flexible communication strategies,
since this model (IC3Net + TarMAC) can learn both when to
communicate and whom to address messages to.

We evaluate IC3Net + TarMAC on the Predator-Prey environment from~\citet{singh_iclr19},
consisting of $n$ predators, with limited vision, moving around
(with a penalty of $r_{\text{explore}}=-0.05$ per timestep)
in search of a stationary prey. Once a predator reaches a prey, it keeps getting positive
reward $r_{\text{prey}}=0.05$ till end of episode \ie till other agents reach prey or
maximum no.~of steps. The prey gets $0.05$ per timestep only till the first
predator reaches it, so it has incentive to not communicate its location.
We compare average no. of steps for
agents to reach the prey during training (\Figref{fig:pp}) and at convergence (\tableref{table:ppnumbers}).
\Figref{fig:pp} shows that using TarMAC with IC3Net leads to significantly faster convergence
than IC3Net alone, and \tableref{table:ppnumbers} shows that TarMAC agents
reach the prey faster.
Results are averaged over $3$ independent runs with different seeds.

%% file: sections/main/conclusions.tex
\section{Conclusions and Future Work}
\label{sec:conclusions}
\vspace{1pt}

We introduced TarMAC, an architecture for multi-agent reinforcement learning that
allows targeted continuous communication between agents via a sender-receiver soft attention
mechanism and multiple rounds of collaborative reasoning. Evaluation on four
diverse environments shows that our model is able to learn intuitive communication
attention behavior and improves performance, even in non-cooperative settings,
with task reward as sole supervision. While TarMAC uses continuous vectors as messages,
it is possible to force these to be discrete, either during training itself
(as in~\citet{foerster_nips16}) or by adding a decoder after learning to ground
these messages into symbols.

In future, we aim to exhaustively benchmark TarMAC on more challenging
$3$D navigation tasks because we believe this is where decentralized targeted
communication is most crucial, as it allows scaling to a large number of agents
with high-dimensional observation spaces. In particular, we are interested
in investigating combinations of TarMAC with recent advances in spatial memory,
planning networks, \etc.

%% file: main.bbl
\begin{thebibliography}{25}
\providecommand{\natexlab}[1]{#1}
\providecommand{\url}[1]{\texttt{#1}}
\expandafter\ifx\csname urlstyle\endcsname\relax
  \providecommand{\doi}[1]{doi: #1}\else
  \providecommand{\doi}{doi: \begingroup \urlstyle{rm}\Url}\fi

\bibitem[Andreas et~al.(2016)Andreas, Rohrbach, Darrell, and
  Klein]{andreas_cvpr16}
Andreas, J., Rohrbach, M., Darrell, T., and Klein, D.
\newblock {Neural Module Networks}.
\newblock In \emph{CVPR}, 2016.

\bibitem[Busoniu et~al.(2008)Busoniu, Babuska, and De~Schutter]{busoniu_08}
Busoniu, L., Babuska, R., and De~Schutter, B.
\newblock {A Comprehensive Survey of Multiagent Reinforcement Learning}.
\newblock \emph{Trans. Sys. Man Cyber Part C}, 2008.

\bibitem[Cho et~al.(2014)Cho, Van~Merri{\"e}nboer, Gulcehre, Bahdanau,
  Bougares, Schwenk, and Bengio]{cho_emnlp14}
Cho, K., Van~Merri{\"e}nboer, B., Gulcehre, C., Bahdanau, D., Bougares, F.,
  Schwenk, H., and Bengio, Y.
\newblock Learning phrase representations using rnn encoder-decoder for
  statistical machine translation.
\newblock In \emph{EMNLP}, 2014.

\bibitem[Das et~al.(2017)Das, Kottur, Moura, Lee, and Batra]{visdial_rl}
Das, A., Kottur, S., Moura, J.~M., Lee, S., and Batra, D.
\newblock {L}earning {C}ooperative {V}isual {D}ialog {A}gents with {D}eep
  {R}einforcement {L}earning.
\newblock In \emph{ICCV}, 2017.

\bibitem[Foerster et~al.(2016)Foerster, Assael, de~Freitas, and
  Whiteson]{foerster_nips16}
Foerster, J., Assael, Y.~M., de~Freitas, N., and Whiteson, S.
\newblock Learning to communicate with deep multi-agent reinforcement learning.
\newblock In \emph{NIPS}, 2016.

\bibitem[Foerster et~al.(2018)Foerster, Farquhar, Afouras, Nardelli, and
  Whiteson]{foerster_aaai18}
Foerster, J., Farquhar, G., Afouras, T., Nardelli, N., and Whiteson, S.
\newblock Counterfactual multi-agent policy gradients.
\newblock In \emph{AAAI}, 2018.

\bibitem[Hausknecht \& Stone(2015)Hausknecht and Stone]{hausknecht_aaai15}
Hausknecht, M. and Stone, P.
\newblock {Deep Recurrent Q-Learning for Partially Observable MDPs}.
\newblock In \emph{AAAI}, 2015.

\bibitem[Hoshen(2017)]{hoshen_nips17}
Hoshen, Y.
\newblock {VAIN}: Attentional multi-agent predictive modeling.
\newblock In \emph{NIPS}, 2017.

\bibitem[Jaderberg et~al.(2018)Jaderberg, Czarnecki, Dunning, Marris, Lever,
  Castaneda, Beattie, Rabinowitz, Morcos, Ruderman, Sonnerat, Green, Deason,
  Leibo, Silver, Hassabis, Kavukcuoglu, and Graepel]{jaderberg_arxiv18}
Jaderberg, M., Czarnecki, W.~M., Dunning, I., Marris, L., Lever, G., Castaneda,
  A.~G., Beattie, C., Rabinowitz, N.~C., Morcos, A.~S., Ruderman, A., Sonnerat,
  N., Green, T., Deason, L., Leibo, J.~Z., Silver, D., Hassabis, D.,
  Kavukcuoglu, K., and Graepel, T.
\newblock Human-level performance in first-person multiplayer games with
  population-based deep reinforcement learning.
\newblock \emph{arXiv preprint arXiv:1807.01281}, 2018.

\bibitem[Jiang \& Lu(2018)Jiang and Lu]{jiang_nips18}
Jiang, J. and Lu, Z.
\newblock Learning attentional communication for multi-agent cooperation.
\newblock \emph{NIPS}, 2018.

\bibitem[Jorge et~al.(2016)Jorge, K{\aa}geb{\"a}ck, and
  Gustavsson]{jorge_iclrw16}
Jorge, E., K{\aa}geb{\"a}ck, M., and Gustavsson, E.
\newblock Learning to play guess who? and inventing a grounded language as a
  consequence.
\newblock In \emph{{NIPS workshop on Deep Reinforcement Learning}}, 2016.

\bibitem[Kottur et~al.(2017)Kottur, Moura, Lee, and Batra]{kottur_emnlp17}
Kottur, S., Moura, J.~M., Lee, S., and Batra, D.
\newblock Natural language does not emerge `naturally' in multi-agent dialog.
\newblock In \emph{EMNLP}, 2017.

\bibitem[Lazaridou et~al.(2017)Lazaridou, Peysakhovich, and
  Baroni]{lazaridou_iclr17}
Lazaridou, A., Peysakhovich, A., and Baroni, M.
\newblock Multi-agent cooperation and the emergence of (natural) language.
\newblock In \emph{ICLR}, 2017.

\bibitem[Lowe et~al.(2017)Lowe, Wu, Tamar, Harb, Abbeel, and
  Mordatch]{lowe_nips17}
Lowe, R., Wu, Y., Tamar, A., Harb, J., Abbeel, P., and Mordatch, I.
\newblock Multi-agent actor-critic for mixed cooperative-competitive
  environments.
\newblock In \emph{NIPS}, 2017.

\bibitem[Mordatch \& Abbeel(2017)Mordatch and Abbeel]{mordatch_arxiv17}
Mordatch, I. and Abbeel, P.
\newblock Emergence of grounded compositional language in multi-agent
  populations.
\newblock \emph{arXiv preprint arXiv:1703.04908}, 2017.

\bibitem[Oliehoek(2012)]{oliehoek_book12}
Oliehoek, F.~A.
\newblock Decentralized {POMDPs}.
\newblock In \emph{Reinforcement Learning: State of the Art}. Springer Berlin
  Heidelberg, 2012.

\bibitem[OpenAI(2018)]{openai_18}
OpenAI.
\newblock {OpenAI Five}.
\newblock \url{https://blog.openai.com/openai-five/}, 2018.

\bibitem[Peng et~al.(2017)Peng, Yuan, Wen, Yang, Tang, Long, and
  Wang]{peng_arxiv17}
Peng, P., Yuan, Q., Wen, Y., Yang, Y., Tang, Z., Long, H., and Wang, J.
\newblock Multiagent bidirectionally-coordinated nets for learning to play
  starcraft combat games.
\newblock \emph{arXiv preprint arXiv:1703.10069}, 2017.

\bibitem[Shoham \& Leyton-Brown(2008)Shoham and Leyton-Brown]{shoham_08}
Shoham, Y. and Leyton-Brown, K.
\newblock \emph{{Multiagent Systems: Algorithmic, Game-Theoretic, and Logical
  Foundations}}.
\newblock Cambridge University Press, 2008.

\bibitem[Singh et~al.(2019)Singh, Jain, and Sukhbaatar]{singh_iclr19}
Singh, A., Jain, T., and Sukhbaatar, S.
\newblock Learning when to communicate at scale in multiagent cooperative and
  competitive tasks.
\newblock In \emph{ICLR}, 2019.

\bibitem[Sukhbaatar et~al.(2016)Sukhbaatar, Szlam, and
  Fergus]{sukhbaatar_nips16}
Sukhbaatar, S., Szlam, A., and Fergus, R.
\newblock Learning multiagent communication with backpropagation.
\newblock In \emph{NIPS}, 2016.

\bibitem[Sutton \& Barto(1998)Sutton and Barto]{sutton_book98}
Sutton, R.~S. and Barto, A.~G.
\newblock \emph{Introduction to Reinforcement Learning}.
\newblock MIT Press, 1998.

\bibitem[Vaswani et~al.(2017)Vaswani, Shazeer, Parmar, Uszkoreit, Jones, Gomez,
  Kaiser, and Polosukhin]{vaswani_nips17}
Vaswani, A., Shazeer, N., Parmar, N., Uszkoreit, J., Jones, L., Gomez, A.~N.,
  Kaiser, {\L}., and Polosukhin, I.
\newblock Attention is all you need.
\newblock In \emph{NIPS}, 2017.

\bibitem[Wu et~al.(2018)Wu, Wu, Gkioxari, and Tian]{house3d}
Wu, Y., Wu, Y., Gkioxari, G., and Tian, Y.
\newblock {Building Generalizable Agents With a Realistic And Rich 3D
  Environment}.
\newblock \emph{arXiv preprint arXiv:1801.02209}, 2018.

\bibitem[Zaheer et~al.(2017)Zaheer, Kottur, Ravanbakhsh, Poczos, Salakhutdinov,
  and Smola]{zaheer_nips17}
Zaheer, M., Kottur, S., Ravanbakhsh, S., Poczos, B., Salakhutdinov, R.~R., and
  Smola, A.~J.
\newblock Deep sets.
\newblock In \emph{NIPS}, 2017.

\end{thebibliography}
